\documentclass[runningheads]{llncs}
\usepackage[T1]{fontenc}
\usepackage{graphicx}
\usepackage{cite}
\usepackage{amsmath,amssymb,amsfonts}
\usepackage{algorithmic}

\usepackage{textcomp}
\usepackage{booktabs}
\usepackage{setspace}
\usepackage{threeparttable}
\usepackage{upgreek}
\usepackage{multirow}
\usepackage{diagbox}
\usepackage{bm}
\usepackage{subfigure}
\usepackage[ruled,linesnumbered]{algorithm2e}

\usepackage[misc]{ifsym}

\usepackage{pifont}
\usepackage{hyperref}
\usepackage{makecell}

\begin{document}
	\title{Feature-prompting GBMSeg: One-Shot Reference Guided Training-Free Prompt Engineering for Glomerular Basement Membrane Segmentation}
	\titlerunning{Feature-prompting GBMSeg}
	%\author{Anonymous}
	\author{Xueyu Liu\inst{1}% 1{Liu, Xueyu}
	 \and
	Guangze Shi\inst{1}% 1{Shi, Guangze}
	 \and
	Rui Wang\inst{2} % 2{Wang, Rui}
	\and
	Yexin Lai\inst{1}% 1{Lai, Yexin}
	 \and 
	Jianan Zhang\inst{1}% 1{Zhang, Jianan}
	 \and
	Lele Sun\inst{1}% 1{Sun, Lele}
	\and
	Quan Yang\inst{1}% 1{Yang, Quan}
	\and
	Yongfei Wu\inst{1}$^{(\textrm{\Letter})}$ %1{Wu, Yongfei} 
	 \and
	Ming Li\inst{1}%1{Li, Ming} 
	 \and
	Weixia Han\inst{3}%3{Han, Weixia} 
	 \and
	Wen Zheng\inst{1}}%1{Zheng, Wen} 
	%%
	%\authorrunning{Anonymous}
	\authorrunning{X. Liu et al.}
	%\institute{Anonymous Organization\\
		%\email{****@*****.**}}
	\institute{Taiyuan University of Technology, Taiyuan, China\\
	\email{wuyongfei@tyut.edu.cn}	
	\and
	University of Science and Technology of China, Hefei, China
	\\
	\and
	Second Hospital of Shanxi Medical University, Taiyuan, China\\}
	\maketitle              % typeset the header of the contribution
	\begin{abstract}
		
		Assessment of the glomerular basement membrane (GBM) in transmission electron microscopy (TEM) is crucial for diagnosing chronic kidney disease (CKD). The lack of domain-independent automatic segmentation tools for the GBM necessitates an AI-based solution to automate the process. In this study, we introduce GBMSeg, a training-free framework designed to automatically segment the GBM in TEM images guided only by a one-shot annotated reference. Specifically, GBMSeg first exploits the robust feature matching capabilities of the pretrained foundation model to generate initial prompt points, then introduces a series of novel automatic prompt engineering techniques across the feature and physical space to optimize the prompt scheme. Finally, GBMSeg employs a class-agnostic foundation segmentation model with the generated prompt scheme to obtain accurate segmentation results. Experimental results on our collected 2538 TEM images confirm that GBMSeg achieves superior segmentation performance with a Dice similarity coefficient (DSC) of 87.27\% using only one labeled reference image in a training-free manner, outperforming recently proposed one-shot or few-shot methods. In summary, GBMSeg introduces a distinctive automatic prompt framework that facilitates robust domain-independent segmentation performance without training, particularly advancing the automatic prompting of foundation segmentation models for medical images. Future work involves automating the thickness measurement of segmented GBM and quantifying pathological indicators, holding significant potential for advancing pathology assessments in clinical applications. The source code will be made available on \href{https://github.com/SnowRain510/GBMSeg}{https://github.com/SnowRain510/GBMSeg}
	\end{abstract}
	

	\section{Introduction}
	\label{sec-intro}
	Renal pathology remains the gold standard for diagnosing various kidney diseases and is essential for formulating treatment strategies and predicting prognosis. Histopathological evaluation of glomerular basement membrane (GBM) plays a crucial role in the diagnosis \cite{fogo2009renal}. Conditions such as membranous nephropathy (MN) and certain primary glomerular diseases can contribute to the increase or decrease in GBM thickness. By precisely evaluating the  GBM, pathologists can provide valuable assistance in determining the specific type of disease affecting a patient. The GBM typically ranges from 100 to 400 nanometers in thickness, requiring transmission electron microscopy (TEM) for accurate visualization of this pathological tissue \cite{zhuo2018alternative}. The abundance of ultrastructures to be examined contributes to a time-consuming and labor-intensive process. Meanwhile, the lack of automated segmentation methods constrains the diagnostic procedure of qualitative analyses of GBM thickness.
	
	Several studies conduct to automate the analysis of TEM images \cite{liu2020pdam,liu2021cleftnet,huang2022semi}. However, successfully executing the entire automated process for GBM segmentation is challenging. The GBM with notably indistinct boundaries with the inner capillary endothelium and outer foot processes, poses a significant challenge for the segmentation of these boundaries \cite{wang2024segmentation}. Recently, some attempts are made to overcome the challenge, M.Rangayyan \textit{et al.} \cite{rangayyan2010segmentation} propose a semi-automatic GBM measurement technique based on the Canny edge detector and active contour method to extract GBM contours under manual supervision. Cao \textit{et al.} \cite{cao2019automatic} introduce a random forest (RF) based machine learning method for the automatic segmentation of basement membranes using 330 annotated TEM images. Wen \textit{et al.} \cite{wen2019semantic} apply the DeepLab-v3-based semantic segmentation algorithm, use the null convolution to expand the perceptual field, control the feature resolution of the image, and achieve a better GBM segmentation using 120 annotated TEM images. Yang \textit{et al.} \cite{yang2022multi} utilize a multi-scale attentional convolutional neural network (CNN) to automatically segment glomerular electron-dense deposits with 1,200 annotated TEM patches. Lin \textit{et al.} \cite{lin2023gclr} tackle self-supervised representation learning to utilize vast unlabeled data and mitigate annotation scarcity, validate on 18,928 unlabeled glomerular TEM images for self-supervised pre-training and fine-tune on 311 labeled images. Wang \textit{et al.} \cite{wang2024segmentation} propose a network architecture, RADS-Net, whose segmentation module combines the advantages of vision transform (ViT) and CNN to achieve better performance in GBM contours segmentation task with 30,000 annotated GBM patches.

	However, those aforementioned computer-aided diagnostic methods for segmenting GBM have several limitations. On the one hand, TEM images exhibit a complex background and high resolution, necessitating significant labor costs for pixel-level annotation. Despite efforts by Wang \textit{et al.} \cite{wang2024segmentation} to simplify the annotation process through semi-automated dataset construction, the specialization of medical data still demands substantial time investment from pathologists for annotation and correction of training data. On the other hand, a domain shift problem arises in TEM images due to different digital devices. Traditional deep learning methods trained and tested for GBM segmentation typically rely on data obtained from the same digital device. This makes the training of the model prone to reaching the local optimum of the current domain, and generalization becomes challenging.
	
	In recent years, research on foundational models in natural language processing (NLP) has been progressively influencing the field of computer vision (CV) \cite{radford2021learning,jia2021scaling}. Especially, Pretrained foundation models (PFMs) represented by DINOv2\cite{oquab2023dinov2} acquire generalized visual features by capturing intricate information at the patch levels, relying solely on raw image data. The learned generic visual features ensure robust zero-shot transferability for downstream tasks. Simultaneously, the Segment Anything Model (SAM) \cite{kirillov2023segment} demonstrates remarkable zero-shot segmentation performance, showcasing considerable potential in open-world image perception. By combining two types of foundational models, a recent work by Liu et al. \cite{liu2023matcher} introduces a new paradigm that implements a training-free segmentation framework on natural images, representing an exploration of automated prompt engineering for the one-shot segmentation task.

	Building upon the aforementioned inspiration and committed to addressing the challenges mentioned above, we present GBMSeg, the training-free model is guided by the one-shot reference image to accurately segment the GBM in TEM images. To enhance the integration of feature matching and SAM for synergistic benefits, we develop a series of automatic prompt engineering techniques across the feature and physical space aimed at improving segmentation quality.  A total of 2538 TEM images from 286 kidney biopsy samples are digitalized as our dataset. GBMSeg achieves the highest Dice Similarity Coefficient (DSC) of 87.27\% in a training-free paradigm using only a single annotated reference image, outperforming the recently proposed one-shot or few-shot segmentation methods.
	
	%\begin{itemize}
	%	\item Using only a one-shot reference image, we construct the model GBMSeg, which can reach the purpose of segmenting the GBM in TEM images without training.
	%	\item We have developed a series of novel automated prompt engineering techniques based on patch-level feature matching, which extracted by DINOv2.
	%	\item We developed a GBM thickness measurement based on the segmentation results, quantifying it as a pathological indicator. Subsequently, a machine learning-based classification model for MN was constructed using both the quantified pathological indicator and clinical indicators. The SHAP values were further employed to analyze and interpret the constructed machine learning model.
	%	
	%\end{itemize}

	\section{Methodology}
	\label{sec-met}
	
\begin{figure*}[htpb]
		\centering
		\centerline{\includegraphics[width=1\linewidth]{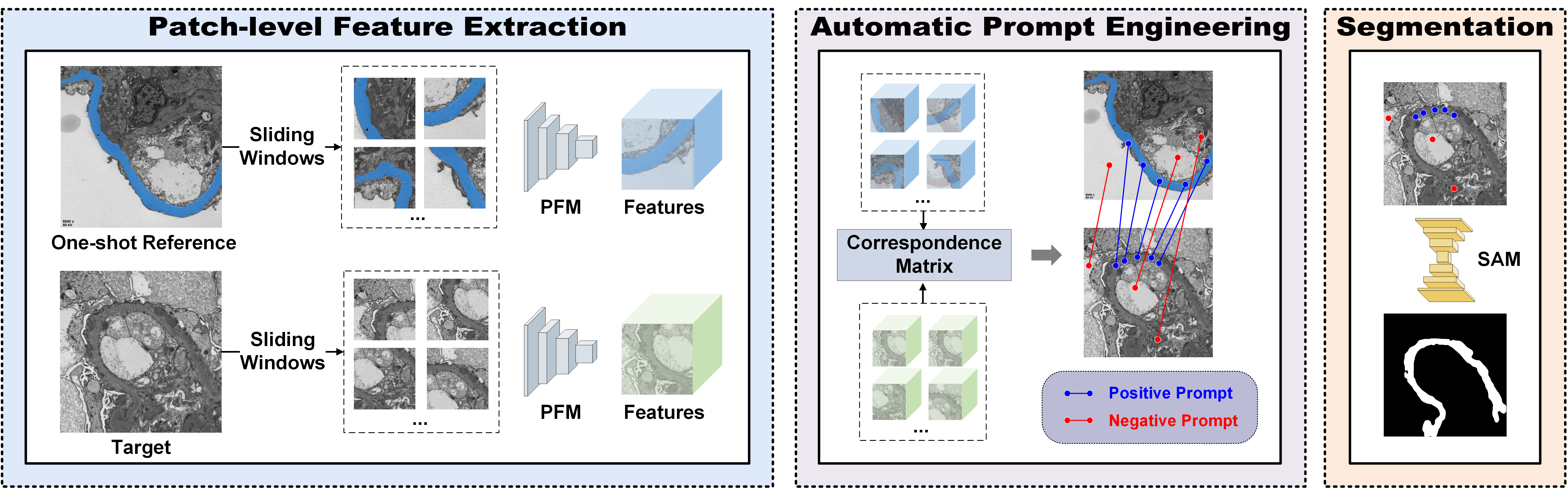}}
		\caption{The workflow of GBMSeg, the one-shot reference guided training-free framework, automates the segmentation of the GBM through three components: Patch-level feature extraction, automatic prompt engineering, and GBM segmentation.}
		\label{stru_all}
		%\vspace{-0.1cm}
	\end{figure*}

	This section introduces our training-free framework, GBMSeg, designed for the segmentation of the GBM in TEM images with a one-shot reference approach. The overview of GBMSeg is illustrated in Fig. \ref{stru_all}. Our framework consists of three components: Patch-level feature extraction, automatic prompt engineering, and GBM segmentation. Specifically, given a target image $x_t$ and a one-shot reference image $x_r$, we divide both into $16 \times 16$ patches $p_t$ and $p_r$ using sliding windows with overlap. Firstly, the Patch-level feature extraction module generates a correspondence matrix $M_s$ by calculating the similarity between $p_t$ and $p_r$. We then utilize the $M_s$ to design a series of prompt engineering for obtaining optimal positive and negative prompt points. Finally, these prompt points are used as inputs to SAM, facilitating the generation of mask proposals. In the following subsections, we will describe the process of automatically generating the prompting scheme in the first two components in detail.
	
	\subsection{Patch-level feature extraction}
	To generate the prompt points of the GBM (or background) in the target image automatically, we need to build a patch-level correspondence matrix between the reference image $x_r$ and the target image $x_t$. Specifically, we first rely on the image encoder from DINOv2\cite{oquab2023dinov2} to extract patch-level features for both $x_t$ and $x_r$ which are represented by $f_t$ and $f_r$, respectively. Patch-level correspondence matrix between the $f_t$ and $f_r$ is computed to discover the most similar regions of the GBM (or background) on the target image. We define a correspondence matrix $M_s$ as follows:
	
	\begin{equation}
		(M_s)_{ij}=\left\lVert f_r^i-f_t^j \right\rVert,
	\end{equation}
	Here, $(M_s)_{ij}$ represents the Euclidean distance between the $i$-th patch features $f_r^i$ from $f_r$ and the $j$-th patch features $f_t^j$ from $f_t$. A smaller value of $(M_s)_{ij}$ indicates that the $j$-th patch is more similar to the $i$-th patch. Thus, we can obtain the patch from the target image that has the highest similarity to each patch in the reference image via $M_s$.

	\subsection{Automatic prompt engineering}
	\label{sec-ape}
\begin{figure}[htpb]
		\centering
		\includegraphics[width=0.70\linewidth]{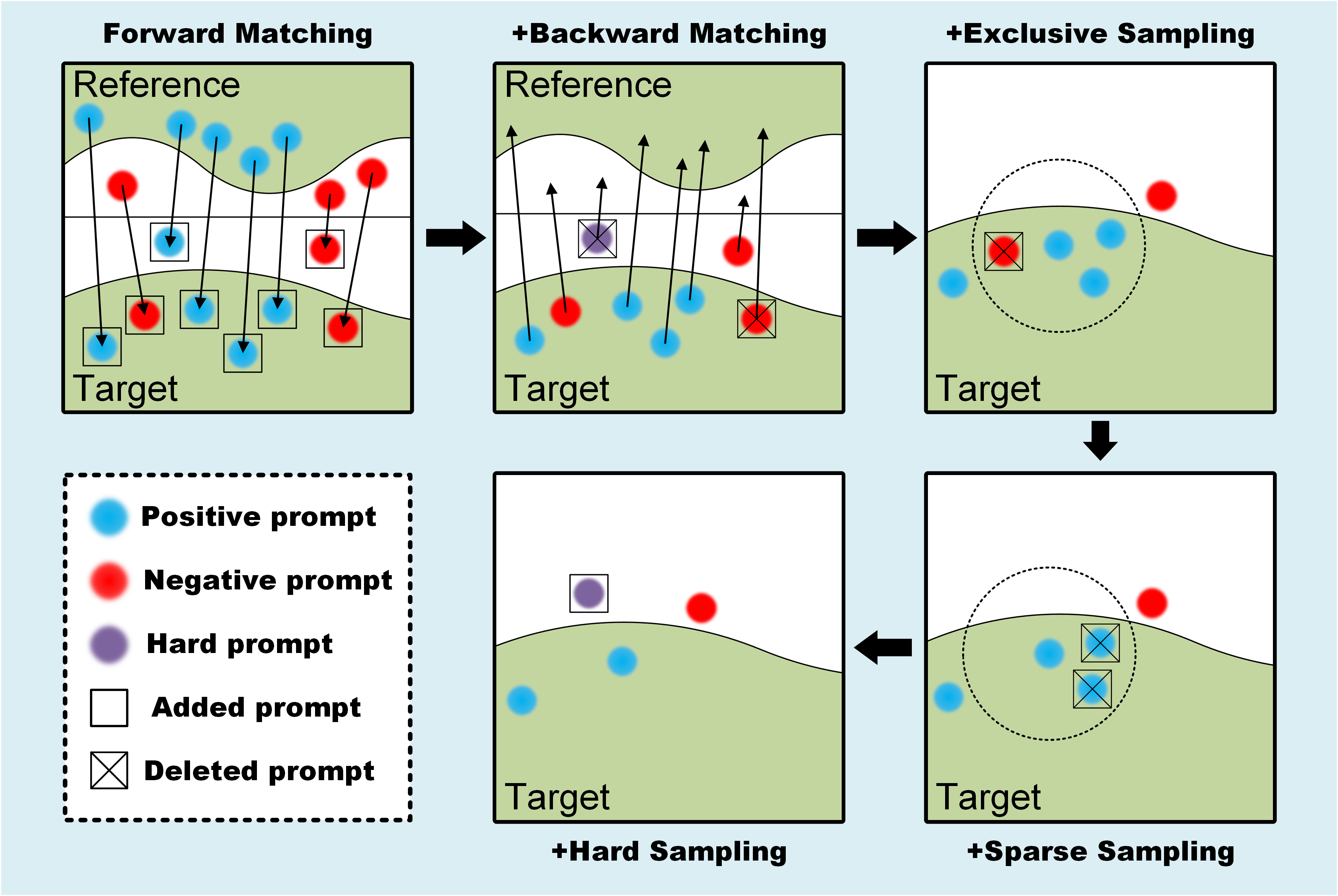}
		\caption{A process for automatic prompt engineering. As the prompt engineering is refined, the prompt scheme is gradually optimized.}
		\label{promptengineering}
	\end{figure}

	SAM serves as a robust foundational model for image segmentation, demonstrating the potential for zero-shot learning through carefully designed prompts. Similar to NLP, the efficacy of prompts profoundly influences SAM's output results. Consequently, leveraging SAM's impressive performance in class-agnostic segmentation, we transform the semantic segmentation task into an endeavor focused on automatically generating high-quality ptompt point. As shown in Fig. \ref{promptengineering}, we devise a series of prompt engineering strategies across the feature and physical space to systematically enhance the segmentation properties of the GBM, a procedural elucidation of which is expounded below.

	\textbf{Forward matching.} Given a target image, our objective is to employ the $M_s$ for the automatic generation of prompt points by selecting the most similar patches from the reference image. We define the patches from $x_r$ whose centers are on the reference mask as $p_r^p$, and the patches whose centers are not on the reference mask as $p_r^n$. For each $x_t$, we can use $M_s$ in the feature space to obtain the most corresponding patch $p_t^p$ (or $p_t^n$) for each $p_r^p$ (or $p_r^n$), and set it center as a positive prompt point (or a negative prompt point).
	
	\textbf{Backward matching.}  Ideally, all patches from the GBM region (or background) in the reference image should be matched as $p_t^p$ (or $p_t^n$) via forward matching. However, in comparison to natural images, the background of TEM images is more complex, and the edges of the GBM are more blurred. Relying solely on forward matching will lead to imprecise and incomplete segmentation results attributed to numerous wrong prompt points. To eliminate incorrect prompt points, we perform backward matching on the prompt points $p_t^p$ (or $p_t^n$) obtained from forward matching to $x_r$ using $M_s$. For each $p_t^p$ (or $p_t^n$), if the most correspond patches are $p_r^p$ (or $p_r^n$), these prompt points are retained. Conversely, if the most corresponding patch is $p_r^n$ (or $p_r^p$), these points are excluded. In addition, these excluded points with correspondence exceeding the mean value would be employed as hard negative sampling prompt points in subsequent steps.
	
	\textbf{Exclusive sampling.} Through forward and backward matching prompt engineering, we have substantiated the accuracy of generated prompt point in the feature space. To further optimize the prompt scheme, we employ exclusive sampling in the physical space. Specifically, by defining a hyperparameter $D_{ex}$, we exclude all negative prompt points within the range defined by each positive prompt point as the center of the circle and $D_{ex}$ as the search radius. This approach ensures the maximization of negative prompt points for errors in the physical space.

	\textbf{Sparse sampling.} The rationale behind our point generation involves iterating over all $p_r^p$ and $p_r^n$ to generate prompt points, resulting in a substantial number of positive prompt points and negative prompt points for each $x_t$. However, the features of background or target exhibit significant diversity. Overemphasizing the same region can cause SAM to disregard other regions. For instance, when the features in the target image strongly correlate with those in the reference image, an abundance of prompt points may concentrate on that region, leading to excessive attention to that specific area and neglecting other regions. Therefore, we employ sparse sampling in the physical space to sparsify positive prompt points (or negative prompt points), aiming for a more balanced focus on the GBM region (or background) to enhance segmentation performance. Specifically, we introduce a hyperparameter $D_{sp}$ to conduct a search with all selected prompt points as the center of the circle and $D_{sp}$ as the search radius. If there exists a set of prompt points of the same class within the range, we compute the average distance between all prompt points and other classes of prompt points separately, retaining the prompt point with the largest distance. This sparsification of prompt points in physical space aims to enhance the subsequent segmentation performance of SAM.

	\textbf{Hard sampling.} To minimize false positive segmentation of background regions similar to GBM regions, we employ a collection of hard negative prompt points. These prompt points are selected based on their correspondence value exceeding the mean value and are removed during the backward matching procedure. These hard negative sampling prompt points, which represent false positive prompt points, are then incorporated into the negative prompt points.

	%\vspace{1.3cm}
	\section{Experimental Results}
	\label{sec-exp}
	\subsection{Dataset preparation}
	
	A total of 286 kidney biopsy samples from patients are collected at the Second Affiliated Hospital of Shanxi Medical University between 2020 and 2022. The dataset comprises a diverse range of chronic kidney diseases, such as membranous nephropathy, diabetic nephropathy, microscopic lesions, and others. From these kidney biopsy samples, a total of 2538 TEM images of glomeruli are digitalized using the JEM-1400 FLASH TEM, operating at 120 kV acceleration voltage, and magnification varied between 2500$\times$ and 15000$\times$. Ethical approval for data collection is obtained from the Ethics Committee on Human Research of the Second Affiliated Hospital of Shanxi Medical University (No. YX.026). All TEM images are labeled by three pathologists using Labelme\cite{russell2008labelme} to annotate the GBM. Only one of these images served as the reference of GBMSeg, while the remaining are employed to evaluate the model's performance.

	\subsection{Implementation Detail}
Our experiments are carried out on a Linux server platform equipped with an NVIDIA Tesla V100. In the patch-level feature extraction stage, we utilize DINOv2 with a ViT-L/14 as the default image encoder. The SAM serves as the segmenter, incorporating ViT-H, ViT-L, and ViT-B, with their performances compared. Notably, our model is training-free, and to emphasize, any training is not employed for segmenting the GBM. During the testing phase, we employ the DSC to evaluate the segmentation performance of the proposed method.
	%\begin{equation}
	%	DSC=\frac{2 \sum\limits_{i=1}^{M} \sum\limits_{c=1}^{C} \left(p_{i, c}\right) \cdot (g_{i, c})}{\sum\limits_{i=1}^{M}\sum\limits_{c=1}^{C} \left(p_{i, c}\right)^{2}+\sum\limits_{i=1}^{M}\sum\limits_{c=1}^{C} (g_{i, c})^{2}}
	% 	\label{Dice}
	%\end{equation}

	\subsection{Ablation study}
	We conduct ablation experiments to evaluate the segmentation performance of different model components. The quantitative segmentation results, compared to various prompt engineering schemes, are presented in Table \ref{result_ablation}. As the components in the automatic prompt engineering continue to be refined, multiple backbone architectures exhibit improved SAM segmentation performance.

\begin{figure}[htpb]
		\centering
		\centerline{\includegraphics[width=1\linewidth]{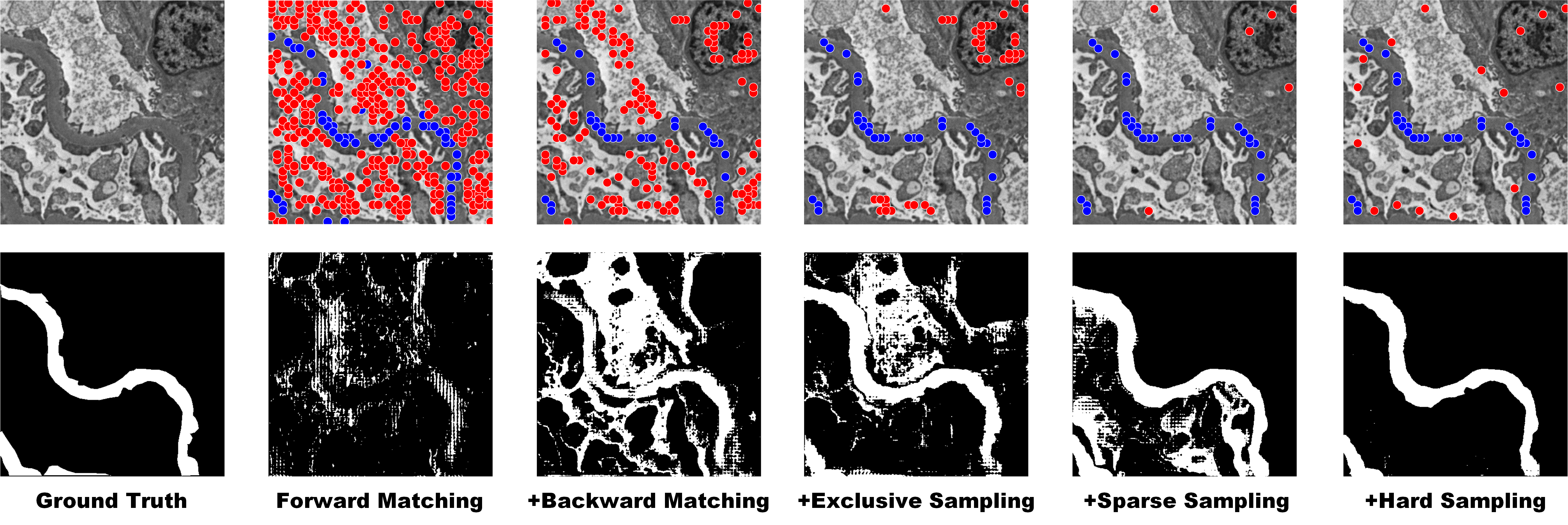}}
		\caption{Different stages of the automatic prompt engineering yield positive and negative prompt schemes, alongside corresponding GBM segmentation results. Notably, as the components of automatic prompt engineering are refined, the segmentation results steadily converge toward the ground truth.}
		\label{perf_ablation}
	\end{figure}

	\begin{table*}[]
		\centering
		\caption{The performance comparison of different model components (Unit: \%).}
		\resizebox{1\linewidth}{!}{
			\begin{tabular}{@{}ccccccccc@{}}
				\toprule
				\makecell{Forward \\ Matching} & \makecell{Backward \\ Matching} & \makecell{Exclusive \\ Sampling} & \makecell{Sparse \\ Sampling} & \makecell{Hard \\ Sampling} & ViT-L & ViT-H & ViT-B&Ave. \\ \midrule
				\ding{51}            &                   &                      &                   &               & 59.73 & 9.68 & 49.77&39.73 \\
				\ding{51}            & \ding{51}             &                      &                   &               & 71.69 & 54.46 & 41.32&55.82 \\
				%\ding{51}            &                   &                      &                   & \ding{51}         & 0.6040 & 0.1147 & 0.5021 \\
				%\ding{51}            &                   & \ding{51}                &                   &               & 0.7932 & 0.7443 & 0.6393 \\
				%\ding{51}            &                   &                      & \ding{51}             &               & 0.8408 & 0.6918 & 0.5722 \\
				%\ding{51}            & \ding{51}             &                      &                   & \ding{51}         & 0.8008 & 0.3090 & 0.5294 \\
				\ding{51}            & \ding{51}             & \ding{51}                &                   &               & 82.11 & 62.38 & 59.50&67.99 \\
				%\ding{51}            & \ding{51}             &                      & \ding{51}             &               & 0.8412 & 0.7092 & 0.6345 \\
				%\ding{51}            &                   & \ding{51}                &                   & \ding{51}         & 0.8269 & 0.6916 & 0.6695 \\
				%\ding{51}            &                   &                      & \ding{51}             & \ding{51}         & 0.8441 & 0.7175 & 0.6989 \\
				%\ding{51}            &                   & \ding{51}                & \ding{51}             &               & 0.6858 & 0.7493 & 0.5879 \\
				%\ding{51}            & \ding{51}             & \ding{51}                &                   & \ding{51}         & 0.8506 & 0.7276 & 0.6602 \\
				%\ding{51}            & \ding{51}             &                      & \ding{51}             & \ding{51}         & 0.8668 & 0.7618 & 0.6971 \\
				\ding{51}            & \ding{51}             & \ding{51}                & \ding{51}             &               & 84.94 & 84.44 & 71.34&80.24 \\
				%\ding{51}            &                   & \ding{51}                & \ding{51}             & \ding{51}         & 0.8453 & 0.7904 & 0.6731 \\
				\ding{51}            & \ding{51}             & \ding{51}                & \ding{51}             & \ding{51}         & \textbf{87.27}& \textbf{85.56} & \textbf{73.77}&\textbf{82.20} \\ \bottomrule
			\end{tabular}
		}
		\label{result_ablation}
	\end{table*}
	 The corresponding qualitative segmentation results are also displayed in Fig. \ref{perf_ablation}. Specifically, forward matching alone leads to a significant number of erroneous and redundant prompt points, resulting in high false positives and false negatives. The introduction of backward matching reduces the occurrence of erroneous prompt points to a certain extent and enhances the recall of segmentation results. Exclusive sampling further corrects for false negative prompt points, contributing to a more comprehensive segmentation of the GBM. Meanwhile, sparse sampling helps refine the prompt points, improving segmentation accuracy by eliminating excessive prompt points in similar regions. Finally, hard sampling is employed to increase the number of challenging negative prompt points, negatively prompting the background area around the target and optimizing segmentation performance.
	
	Additionally, we conducted hyperparameter selection experiments for exclusive sampling and sparse sampling. For exclusive sampling, the performance is optimal when $D_{ex}$ selects 25\% of the image size. For sparse sampling, the best performance is achieved when $D_{sp}$ for positive sampling points is set to 0, and $D_{sp}$ for negative sampling points is set to 12.5\% of the image size. This may be due to the fact that a dense distribution of negative prompts in regions with heterogeneous background features can degrade SAM segmentation performance. Conversely, in targets with homogeneous features, a dense positive prompt distribution does not degrade SAM segmentation performance and helps counteract the impact of some erroneous negative prompts.
	
	\subsection{Performance comparison with few-shot and one-shot methods}

To assess the effectiveness of the synergy between the proposed automatic prompt engineering and SAM, we compare GBMSeg with state-of-the-art few-shot or one-shot segmentation networks \cite{min2021hypercorrelation, hong2022cost}, as well as recently introduced training-free segmentation frameworks \cite{wang2023seggpt, zhang2023personalize, liu2023matcher}. The experimental results, presented in Table \ref{result_com}, demonstrate that GBMSeg not only outperforms the current state-of-the-art training-free methods but also excels over both the one-shot and few-shot training methods. Our model achieves optimal performance with minimal training resources by implementing an effective prompt scheme to assist the SAM in segmenting the GBM.
	
	\begin{table}[]
		\caption{The performance comparison with few-shot and one-shot methods (Unit: \%).}
		\centering
		\begin{tabular*}{\hsize}{@{}@{\extracolsep{\fill}}lcccc@{}}
			\toprule
			Methods          &Annotated samples&SAM-based& Traing-free & DSC      \\ \midrule
			HSNet-1  \cite{min2021hypercorrelation}  &One-shot       & \ding{55}   & \ding{55}      & 21.03  \\
			HSNet-5 \cite{min2021hypercorrelation}   &Five-shot        & \ding{55}   & \ding{55}         & 50.34 \\
			VAT-1  \cite{hong2022cost}      &One-shot        & \ding{55}    & \ding{55}         & 68.74 \\
			VAT-5  \cite{hong2022cost} &Five-shot & \ding{55}  &\ding{55}         & 78.81\\
			SegGPT\cite{wang2023seggpt}    &One-shot       & \ding{55}   & \ding{51}       & 74.31  \\
			PerSAM\cite{zhang2023personalize}    &One-shot       & \ding{51}    & \ding{51}       &42.39          \\
			Matcher\cite{liu2023matcher}    &One-shot       & \ding{51}   & \ding{51}       &69.09 \\
			GBMSeg (Our work)   &One-shot   & \ding{51}    & \ding{51}    &    \textbf{87.27}       \\ \bottomrule
		\end{tabular*}
		\label{result_com}
	\end{table}

	\section{Conclusion}
	\label{sec-Conclusion}
	
	In this study, we introduce GBMSeg, a training-free model designed to segment the GBM in TEM images, guided only by a one-shot reference. The proposed framework leverages the robust feature matching capabilities inherent in a universal feature extraction model to generate initial positive and negative prompt points. Subsequently, it employs a series of novel automatic prompt engineering techniques across the feature and physical space to obtain optimized prompt schemes. The refined prompt scheme is then input into a foundation segmentation model, resulting in the final segmentation outcome. GBMSeg demonstrates exceptional performance, achieving a DSC of up to 87.27\% on 2538 glomerular TEM images, using only one annotated reference image. This performance establishes GBMSeg as outperforming the recently proposed one-shot or few-shot segmentation methods. Rigorous experiments on ablation studies also substantiate the efficacy of each component in the automatic prompt engineering process. In summary, GBMSeg adeptly and efficiently segments GBM only using a one-shot reference, offering a training-free paradigm.  Future endeavors include the quantification of pathological metrics hold significant potential for enhancing pathological assessment and decision-making in clinical applications.

	%\section*{Acknowledgment}
	%This work was supported by  the National Natural Science Foundation of China [Grant No. 61901292], and the National Key Technology R\&D Program of China [Grant No. 2018YFB1701700], and Shanxi Provincial Graduate Education Innovation Project [Grant No.2022Y196]
	
	%\section*{Acknowledgment}
	%The authors sincerely acknowledge the pathologists who offering great assistance. We also sincerely acknowledge the editor and any reviewers for their valuable advice.
	
\subsubsection{\discintname}
The authors have no competing interests to declare that are relevant to the content of this article.
	\bibliographystyle{splncs04}
	\bibliography{gol_bib}
\end{document}

\title{Feature-prompting GBMSeg: One Shot Reference Guided Training-Free Prompt Engineering for Glomerular Basement Membrane Segmentation}
\titlerunning{Feature-prompting GBMSeg}
\author{Anonymous}
%\author{Xueyu Liu\inst{1} \and
%Guangze Shi\inst{1} \and
%Rui Wang\inst{2} \and
%Yexin Lai\inst{1} \and
%Jianan Zhang\inst{1} \and
%Yongfei Wu\inst{1}$^{(\textrm{\Letter})}$ \and
%Ming Li\inst{1} \and
%Weixia Han\inst{3} \and
%Wen Zheng\inst{1}}
%%
\authorrunning{Anonymous}
%\authorrunning{X. Liu et al.}
\institute{Anonymous Organization\\
	\email{****@*****.**}}
%\institute{Taiyuan University of Technology, Taiyuan, China
%\email{wuyongfei@tyut.edu.cn}	
%\and
%University of Science and Technology of China, Hefei, China 
%\\
%\and
%Second Hospital of Shanxi Medical University, Taiyuan, China\\}
%%
\maketitle

\section{Supplementary Experimental Results}

\subsection{Performance analysis of hyperparameter selection}

\subsubsection{Hyperparameter selection of exclusive sampling}

In independence sampling, our ideal scenario involves having no negative prompt points within the segmentation target. To achieve this, we implement negative prompt point removal with $D_{ind}$ as the search radius, using each positive sample point as the center of the search circle. Consequently, the choice of the hyperparameter $D_{ind}$ significantly influences the model's performance. As we illustrated in the Table \ref{result_ind}, we assign $D_{ind}$ a value corresponding to different proportions of the image size, resulting in various prompt point schemes. Additionally, we conduct segmentation experiments with multiple backbone SAMs, revealing that optimal performance is achieved when $D_{ind}$ ranges from 0 to 25.00 percent of the image size, yielding an average DSC of 0.8220. Conversely, performance diminishes as $D_{ind}$ increases to 50.00 percent of the image size.

\begin{table}[htpb]
	\centering
	\caption{The performance comparison of different $D_{ind}$}
	\begin{tabular}{@{}ccccc@{}}
		
		\toprule
		$D_{ind}$ & ViT-L & ViT-H & ViT-B&Ave. \\ \midrule
		50.00\%   & 0.8528 & 0.8241 & 0.7231&0.8000 \\
		25.00\%& \textbf{0.8727}& \textbf{0.8556} & \textbf{0.7377}&\textbf{0.8220} \\
		12.50\%    & 0.8702 & 0.8002 & 0.7147&0.7950 \\
		0     & 0.8668 & 0.7618 & 0.6971&0.7752 \\ \bottomrule
	\end{tabular}
	\label{result_ind}
\end{table}

\subsubsection{Hyperparameter selection of sparse sampling}
\label{sec-dis-sam}
Excessive prompting for similar regions can influence the segmentation performance of SAM. When the search radius $D_{dis}$ is increased, the distribution of prompt points becomes sparser, while a smaller $D_{dis}$ leads to a denser distribution. As illustrated in Table \ref{result_dis}, we assign distinct values to $D_{dis}^p$ (or $D_{dis}^n$) based on different proportions of the image size, resulting in various positive (or negative) prompt point schemes, respectively. In the positive prompt points, the model performance consistently decreases as $D_{dis}^p$ increases. Conversely, for negative prompt points, the model performance reaches its peak when $D_{dis}^n$ increases to 12.50 percent of the image size. The diversity of background features can lead to a deterioration of SAM segmentation performance when a dense distribution of negative prompts is employed. Conversely, in segmentation targets characterized by more homogeneous features, a dense positive prompt distribution proves effective in ensuring the integrity of the segmentation.

\begin{table}[htpb]
	\centering
	\caption{The performance comparison of different $D_{dis}$}
	\begin{tabular}{@{}cccccc@{}}
		\toprule
		$D_{dis}^p$ & $D_{dis}^n$& ViT-L & ViT-H & ViT-B& Ave. \\ \midrule
		0    & 0    & 0.8506 & 0.6576 & 0.6302 &0.7128\\
		6.25\%   & 0    & 0.8248 & 0.5643 & 0.5060&0.6317 \\
		12.50\%   & 0    & 0.7743 & 0.3630 & 0.4242&0.5205 \\
		25.00\%  & 0    & 0.5986 & 0.2012 & 0.1619&0.3206 \\
		0    & 6.25\%   & 0.8696 & 0.7894 & 0.7132 &0.7907\\
		0    & 12.50\%  & \textbf{0.8727}& \textbf{0.8556} & \textbf{0.7377}&\textbf{0.8220} \\
		0    & 25.00\%  & 0.8513 & 0.8258 & 0.7233& 0.8001\\ \bottomrule	
	\end{tabular}
	\label{result_dis}
\end{table}

\subsubsection{Negative prompt selection}
\label{sec-neg-sel}
In our prompt engineering, we designate background prompts for the prompting points derived from the background, and hard prompts for the prompting points obtained from target deletion and re-screening. We conducted experiments to examine the impact of various types of negative sampling points on segmentation results. As illustrated in the table, the segmentation performance, when both types of prompts are utilized as negative prompts, surpasses the performance achieved when using negative prompts or hard prompts alone.
\begin{table}[htpb]
	\centering
	\caption{The performance comparison of different negative prompt}
	\begin{tabular}{@{}ccccc@{}}
		\toprule
		Negative prompt  & ViT-L & ViT-H & ViT-B &Ave.\\ \midrule
		Background only      & 0.8494 & 0.8444 & 0.7134&0.8024\\
		Hard only     & 0.7724 & 0.8034 & 0.7293&0.7684 \\
		Background w Hard & \textbf{0.8727}& \textbf{0.8556} & \textbf{0.7377}&\textbf{0.8220}\\ \bottomrule
	\end{tabular}
	\label{result_hard}
\end{table}